\documentclass{article}

\usepackage{arxiv}

\usepackage[utf8]{inputenc} 
\usepackage[T1]{fontenc}    
\usepackage{hyperref}       
\usepackage{url}            
\usepackage{booktabs}       
\usepackage{amsfonts}       
\usepackage{nicefrac}       
\usepackage{microtype}      
\usepackage{lipsum}		
\usepackage{graphicx}
\usepackage{natbib}
\usepackage{doi}
\usepackage{times}
\usepackage{latexsym}
\usepackage{graphicx}
\usepackage{float}
\usepackage{amsmath}
\usepackage{marvosym}
\usepackage{lineno}

\usepackage{multirow}
\usepackage{amssymb}

\title{Cross Domain Few-Shot Learning via Meta Adversarial Training}

\author{
  \textbf{Jirui Qi} \\
  BDBC and SKLSDE \\
  Beihang University, China \\
  \texttt{\footnotesize qijr@act.buaa.edu.cn} \\\\
  \textbf{Chune Li} \\
  BDBC and SKLSDE \\
  Beihang University, China \\
  \texttt{\footnotesize lice@act.buaa.edu.cn} \\\And
  \textbf{Richong Zhang\textsuperscript{\Letter}} \\
  BDBC and SKLSDE \\
  Beihang University, China \\
  \texttt{\footnotesize zhangrc@act.buaa.edu.cn} \\\\
  \textbf{Yongyi Mao} \\
  School of EECS \\
  University of Ottawa, Canada \\
  \texttt{\footnotesize ymao@uottawa.ca} \\
  }

\hypersetup{
pdftitle={Cross Domain Few-Shot Learning via Meta Adversarial Training},
pdfsubject={},
pdfauthor={Jirui Qi, Richong Zhang, Chune Li, Yongyi Mao},
pdfkeywords={Meta-learning, Cross Domain, Domain Adaptation, FewRel 2.0},
}

\begin{document}
\maketitle
\begin{abstract}
Few-shot relation classification (RC) is one of the critical problems in machine learning. Current research merely focuses on the set-ups that both training and testing are from the same domain. However, in practice, this assumption is not always guaranteed. In this study, we present a novel model that takes into consideration the afore-mentioned cross-domain situation. Not like previous models, we only use the source domain data to train the prototypical networks and test the model on target domain data. A meta-based adversarial training framework (\textbf{MBATF}) is proposed to fine-tune the trained networks for adapting to data from the target domain. Empirical studies confirm the effectiveness of the proposed model.
\end{abstract}


\section{Introduction}
Previous research on few-shot relation classification(RC) only considers the situation that the datasets for meta-training and meta-testing are from the same domain until the presentation of FewRel 2.0 \citep{gao-etal-2019-fewrel} dataset, whose meta-training and meta-testing differ vastly from each other:
the meta-training data derives from Wikidata, a comprehensive and non-professional dataset,
while meta-testing data is from Pubmed, a medical dataset greatly different from Wikidata in morphology and syntax,
which leads to a novel cross-domain few-shot learning problem.

Some previous works adopt adversarial training
to solve the cross-domain problems, like Prototypical-ADV model \citep{gao-etal-2019-fewrel}, which introduces a adversarial training method with a number of unlabeled target-domain data to make the encoded target-domain and source-domain instances more similar. Moreover, on the basis of it, (\citealp{DBLP:journals/corr/abs-2006-12816}) further uses the unlabeled target-domain data. The unlabeled target-domain data gets pseudo-labels generated by cluster miner, and is used for meta-training.

In these previous works, they use additional unlabeled target-domain (meta-testing) instances
to do the adversarial training in the meta-training process.
However, in real life, it is extraordinary difficult for us to get additional data in the target area, even it is unlabeled. 
In other words, we only have the small amount target-domain data in the support set of each few-shot task, usually 5 to 50.
On the other hand, we can have a large amount of source domain data to adversarial train a robust encoder.

For the above consideration, in this paper we propose a meta-based adversarial training framework (\textbf{MBATF}) for cross-domain few-shot learning.
Instead of instance-level features, we pay more attention to task-level features and relation-level features.
From the task level, we propose a Meta-ADV module, only using the source-domain instances, to make the encoded meta-training vectors and the encoded meta-testing vectors closer to each other, which is significantly different from the existing work. From the relation level, we adopt a relation-scoring module to figure out the representative characteristics of each relation, since although there is a big difference between the source domain and the target domain, some similarities still exist. Therefore, we can highlight these characteristics in order to increase the similarities between the meta-training vectors and the meta-testing vectors.

In the actual scenario, we conduct experiments on a large-scale cross-domain dataset FewRel 2.0 
and a slight-scale cross-domain dataset FewRel 1.0
to demonstrate the effectiveness of our model.
Meanwhile, we compare our model performance with the models that use additional target domain data. The results show that the performance of our models is similar to theirs, and some are even better than their models,
which further proves the advantages of our model.

\begin{figure*}
    \centering
    \includegraphics[width=1.0\linewidth]{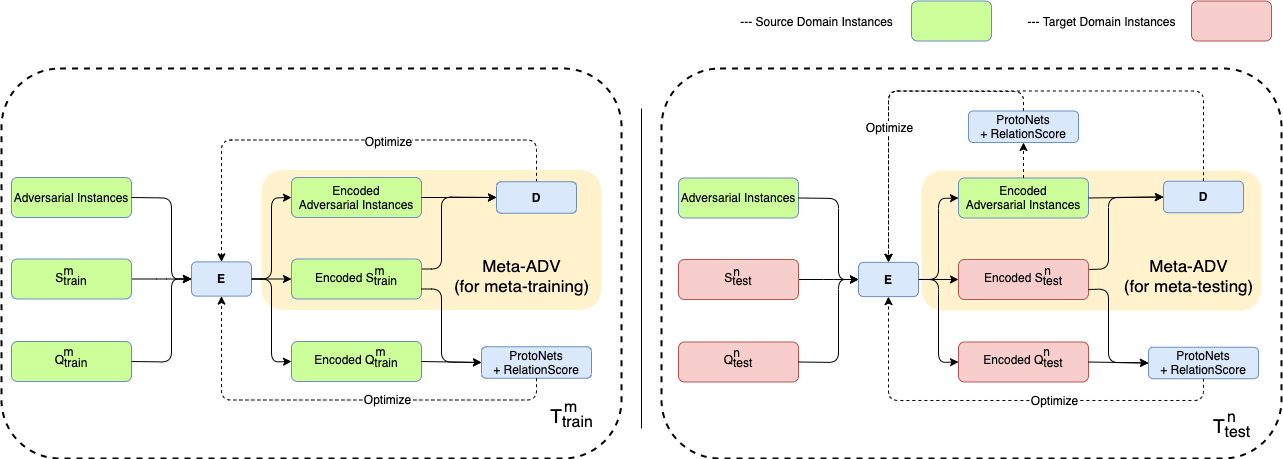}
    \caption{MBATF model structure.}
    \label{model-structure}
\end{figure*}

\section{Method}
In this section, we describe the overall model framework (MBATF) for few-shot relation classification.
After a quick review of prototypical network approach for few-shot relation classification problem, 
we propose our meta-based adversarial training (Meta-ADV) approach for cross-domain few-shot learning, 
And then we describe our relation-level scoring module (Relation-scoring).
The MBATF overall model is shown in Fig.\ref{model-structure}.

\subsection{Prototypical Network for Few-shot Relation Classification}
We focus on the few-shot relation classification problem, although the proposed framework can apply to any other few-shot text classification problems.
Given a sentence of a sequence of words
$s=(w_1,w_2,\cdots)$, and two entities $(h,t)$ identified as token positions in the sentence, the goal of relation classification is to figure out a semantic relation $r\in \mathbf{R}$ between the two entities, where $\mathbf{R}$ is a predefined relation set.

We consider the typical $N$-way $K$-shot few-shot classification scenario,
where tasks are divided into
meta-training tasks
and meta-testing tasks.
Each meta-task involves a $N$-way  classification.
Each meta-task $\mathbf{T}$ has a support set $\mathbf{S}$ and a query set $\mathbf{Q}$, both with $K$ 
instances for each of $N$ classes.
The goal is to classify instances in any meta-testing query set $\mathbf{Q}_{\rm test}^i$  .
\begin{eqnarray}
\mathbf{M}_{\rm train}&=&\{ \mathbf{T}_{\rm train}^1, \mathbf{T}_{\rm train}^2,\cdots,\mathbf{T}_{\rm train}^m\} \\
\mathbf{M}_{\rm test}&=&\{ \mathbf{T}_{\rm test}^1, \mathbf{T}_{\rm test}^2,\cdots,\mathbf{T}_{\rm test}^n\}\\
\mathbf{T}_{\rm train}^i&=&(\mathbf{S}_{\rm train}^i,\mathbf{Q}_{\rm train}^i), i=1,\cdots,m\\
 \mathbf{T}_{\rm test}^i&=&(\mathbf{S}_{\rm test}^i,\mathbf{Q}_{\rm test}^i),i=1,\cdots,n
\end{eqnarray}


In this paper, we focus on prototypical network (\citealp{DBLP:conf/nips/SnellSZ17}) meta-learning approach,
although the proposed framework can apply to any other metric-based meta-learning approaches including Relation Networks \citep{DBLP:conf/cvpr/SungYZXTH18} and Siamese Neural Networks \citep{Koch2015SiameseNN}, etc.
For each meta-task, the prototypical network first calculates a prototype $\mathbf{c}_r$ for each class based on the support set, and then predicts labels for query instance by comparing with the distance between the query and each class prototype.
\begin{eqnarray}
\mathbf{c}_r=\frac{1}{|\mathbf{S}_r|}\sum_{(x_i,y_i)\in \mathbf{S}_r }  \mathbf{E}(x_i), \ \ r=1,\cdots,N \\
p(\hat{y}=r|x)\! = \!
\frac{\exp(-d( \mathbf{E}(x), \mathbf{c}_r ))}
{\sum_{j=1}^{N} \exp(-d(\mathbf{E}(x),\mathbf{c}_j))},
x\in \mathbf{Q}
\end{eqnarray}
where $\mathbf{S}_r$ are the support instances of relation $r$,
$\mathbf{E}(x)$ is the encoding vector of an instance $x=(s,h,t)$
got from an encoder $\mathbf{E}$. 
$d(\cdot,\cdot)$ is typically the Euclidean distance.

In the meta-training process,
the prototypical network makes
predictions for each $\mathbf{Q}_{\rm train}$ based on $\mathbf{S}_{\rm train}$ .
By minimizing these prediction errors,
the model learns a good encoder $\mathbf{E}$.
With this learned encoder,
the prototypical network can make
predictions for any $\mathbf{Q}_{\rm test}$ based on $\mathbf{S}_{\rm test}$ in the meta-testing process.

\subsection{Meta-based Adversarial Training}
We propose a meta-based adversarial training (Meta-ADV) approach to improve the domain generalization ability of meta-learning models. Different from existing work, we only use the source-domain (meta-training) instances for adversarial training in the meta-training process. Only in the meta-testing process, we use the N-way K-shot target support instances to finally finetune and adversarial training the robust encoder.
For each meta-task, besides the support set $\mathbf{S}$ and the query set $\mathbf{Q}$,
we also construct a adversarial set $\mathbf{A}$ from the source domain.
$\mathbf{A}$ consists of $K$ instances from $N$ relations
which are randomly sampled from source domain excluding current meta-task relations.

More specifically, in the meta-training process, the prototypical network makes predictions for $\mathbf{Q}_{\rm train}$ based on $\mathbf{S}_{\rm train}$ and update the encoder $\mathbf{E}(x)$ based on the predictions.
Besides, there is a discriminator $\mathbf{D}$ which updates itself trying to distinguish $\mathbf{S}_{\rm train}$ and $\mathbf{A}$,
At the same time, the encoder updates itself to fool the discriminator.
\begin{eqnarray}
&\min_{\theta_\mathbf{E}} &  \sum_{(x,y) \in \mathbf{Q}_{\rm train} }\!\!l( p(\cdot|x),y) \\
&\min_{\theta_\mathbf{D}} & \sum_{x\in \mathbf{S}_{\rm train} \cup \mathbf{A}}\!\!l(\mathbf{D}(\mathbf{E}(x)),\mathbb{I}_{\mathbf{S}_{\rm train}}(x)) \\
&\min_{\theta_\mathbf{E}} & \sum_{x\in \mathbf{S}_{\rm train} \cup \mathbf{A}} \!\!\!\!\!\!\!\!l(\mathbf{D}(\mathbf{E}(x)),\!1\!-\!\mathbb{I}_{\mathbf{S}_{\rm train}}\!(x))
\end{eqnarray}
where $\theta_\mathbf{E}$, $\theta_\mathbf{D}$ denote the parameters of the encoder and discriminator respectively,
$l(\cdot,\cdot)$ denotes the cross entropy loss between two distributions,
$\mathbb{I}_\mathbf{S}(x)$ is the indicator function, which takes value 1 when $x\in \mathbf{S}$ otherwise takes value 0.

In the meta-testing process, we do the Meta-ADV with $\mathbf{S}_{\rm test}$ from the target domain, while $\mathbf{A}$ is still from source domain.
We also finetune the encoder based on $A$ when doing Meta-ADV.
\begin{eqnarray}
&\min_{\theta_\mathbf{E}} &\sum_{(x,y) \in \mathbf{A} }l( p(\cdot|x),y) \\
&\min_{\theta_\mathbf{D}} &\sum_{x\in \mathbf{S}_{\rm test} \cup \mathbf{A}}l(\mathbf{D}(\mathbf{E}(x)),\mathbb{I}_{\mathbf{S}_{\rm test}}(x)) \\
&\min_{\theta_\mathbf{E}} &\sum_{x\in \mathbf{S}_{\rm test} \cup \mathbf{A}} \!\!\!\!\!\!l(\mathbf{D}(\mathbf{E}(x)),\!1\!-\mathbb{I}_{\mathbf{S}_{\rm test}}\!(x))
\end{eqnarray}
After the Meta-ADV,
we calculate the prototypes based on $\mathbf{S}_{\rm test}$ and encoder to make final prediction for $\mathbf{Q}_{\rm test}$.

\subsection{Relation-level Scoring}
Besides, we also introduce a relation-level scoring module $\mathbf{G}$ in the prototype networks in our MBATF.
As also noted in (\citealp{DBLP:conf/aaai/GaoH0S19}), some dimension are more discriminative for classifying special relations in the feature space,
the scoring module helps to show which dimension of the encoded instances in this relation is more representative and which is less.
For each meta-task, the encoded embedding of $k$ support instances of relation $r$ are concatenated as a $K\times d$ feature map, where $d$ is the embedding dimension. And a CNN network translates it to a score vector $g_r\in R^d$. This score vector is applied to the Euclidean distance calculation in the prototype networks as follows.
\begin{eqnarray}
g_r &=& \mathbf{G}(  \{ {\mathbf E}(x),x \in \mathbf{S}_r \} ) \\
d( \mathbf{E}(x), \mathbf{c}_r )&=&g_r  (  \mathbf{E}(x)- \mathbf{c}_r )^2
\end{eqnarray}

\section{Experiments}

\subsection{Dataset and Baseline}
Our experiments are composed of two parts.
In the first experiment, we versify the effectiveness of our meta-based adversarial model MBATF on FewRel 2.0 (Wiki$\to$Pubmed)\citep{gao-etal-2019-fewrel} and FewRel 1.0 (Wiki$\to$Wiki)\citep{han-etal-2018-fewrel}.
We compare with the prototypical networks as baseline as introduced on FewRel 2.0 benchmark.

The second experiment compares our model with Prototypical-ADV\citep{gao-etal-2019-fewrel}.
Note that the comparison is largely bias to Prototypical-ADV,
since the Prototypical-ADV method has used unlabeled target-domain instances in meta-training,
while we did not.
To our best knowledge, in the cross-domain few-shot learning, there is no existing adversarial training approach without using any target-domain data as our MBATF did.
Besides FewRel 2.0/1.0, we compare with Prototypical-ADV
on another two new set-ups: Wikidata$\to$SemEval-2010 Task 8 dataset \citep{DBLP:conf/naacl/HendrickxKKNSPP09}, and Wikidata$\to$NYT-10 \citep{DBLP:conf/pkdd/RiedelYM10}.

\subsection{Experimental Settings}
This section we describe the experimental settings.
For the encoder, we use Glove+CNN model.
First, we exploit the Glove pre-trained model \citep{pennington-etal-2014-glove}
and embedding the words to 50-dimension vectors.
Then both the word embedding and position embedding are passed to a CNN encoder, which follows the settings introduced in (\citealp{DBLP:conf/coling/ZengLLZZ14}).
The encoder finally encodes each instance to a vector of length 230.
We exploit a 2-layer MLP discriminator with 230 hidden dimension.
The relation-level scoring module is a 3-layer CNN with  32,64 and 1 output channels respectively.

We use the stochastic gradient descent algorithm optimize the encoder, discriminator and relation scoring modules with the same learning rate 0.1.
In order to highlight the impact of limited target domain data, compared with only 1 iteration ADV in each meta-task in the meta-training process, 5 ADV iterations were conducted in each meta-task in the meta-testing process.



\subsection{Evaluation Results}

\subsubsection{Results on FewRel 2.0 and FewRel 1.0}
Table \ref{tb:fewrel} shows the results of our model on FewRel 2.0 and FewRel 1.0.
As an ablation study, we first test the ProtoNets+MetaADV model, i.e. the baseline ProtoNets incorporated with our proposed MetaADV module, to show the effectiveness of our meta-based adversarial training method.
In the 10-way 5-shot FewRel 2.0 task, the improvement on accuracy is large, from 37.70\% to 40.64\%.
Secondly, we import the relation-level scoring module and the results shows that the full model, with both MetaADV module and
relation-scoring module, is better than the baseline model on all the few-shot experiments shown in the table, especially larger improvement on cross-domain FewRel 2.0.
The experimental results demonstrate the effectiveness of our proposed model in cross-domain few-shot learning.

\begin{table*}
\centering
\begin{tabular}{|c|c|c|c|c|c|}
\hline
\multirow{2}{*}{Datasets} &
\multirow{2}{*}{Model} &
\multirow{2}{*}{5 way 1 shot} &
\multirow{2}{*}{5 way 5 shot} &
\multirow{2}{*}{10 way 1 shot} &
\multirow{2}{*}{10 way 5 shot} 
\\
 & & & & & \\

\hline
\multirow{3}{*}{\shortstack{FewRel 2.0\\ (Wiki$\to$Pubmed)}}
& {ProtoNets} & {33.69} & {48.71} & {22.02} & {37.70}  \\
& {ProtoNets+MetaADV} & {34.58} & {49.15} & {22.95} & {40.64}  \\
& {MBATF} & \textbf{37.06} & \textbf{55.78} & \textbf{23.75} & \textbf{40.82}  \\
\hline
\multirow{3}{*}{\shortstack{FewRel 1.0\\ (Wiki$\to$Wiki)}}
& {ProtoNets} & {71.05} & {88.41} & {60.94} & {80.45} \\
& {ProtoNets+MetaADV} & {72.02} & {88.96} & {61.19} & {80.97}  \\
& {MBATF} & \textbf{72.48} & \textbf{89.36} & \textbf{62.77} & \textbf{81.95} \\
\hline
\end{tabular}
\caption{\label{tb:fewrel} Performance of our MBATF (ProtoNets+MetaADV+RelationScore) on FewRel 2.0 and FewRel 1.0.}
\end{table*}

\subsubsection{Comparison with Prototypical-ADV}
Meanwhile, we compare our model with Prototypical-ADV model which uses extra target-domain instances. The results in Table \ref{Performance-2} indicate that our performance is similar to its, and even better under some testing set-ups, especially on FewRel 1.0 (Wiki $\to$ Wiki) and Wiki $\to$ NYT.

\begin{table*}
\centering
\begin{tabular}{|c|c|c|c|c|c|}
\hline
\multirow{2}{*}{\shortstack{Datasets\\(Source$\to$Target)}} &
\multirow{2}{*}{Model}&
\multicolumn{4}{c|}{Tasks} \\
\cline{3-6}
&   & 5 way 1 shot & 5 way 5 shot & 10 way 1 shot & 10 way 5 shot\\
\hline

\multirow{2}{*}{\shortstack{FewRel 2.0\\(Wiki$\to$Pubmed)}}
& {Prototypical-ADV} & \textbf{37.45} & \textbf{56.58} & \textbf{26.70} & \textbf{43.94}   \\
& {MBATF} & {37.06} & {55.78} & {23.75} & {40.82} \\
\hline

\multirow{2}{*}{\shortstack{FewRel 1.0\\(Wiki$\to$Wiki)}}
& {Prototypical-ADV} & {66.57} & {83.39} & {54.29} & {76.60}   \\
& {MBATF} & \textbf{72.48} & \textbf{89.36} & \textbf{62.77} & \textbf{81.95} \\
\hline

\multirow{2}{*}{Wiki$\to$SemEval}
& {Prototypical-ADV} & \textbf{34.01} & {46.89} & \textbf{22.07} & {35.42}   \\
& {MBATF} & {32.46} & \textbf{52.32} & {21.09} & \textbf{40.48} \\
\hline

\multirow{2}{*}{Wiki$\to$NYT}
& {Prototypical-ADV} & {62.48} & {81.81} & {51.61} & {73.26}   \\
& {MBATF} & \textbf{67.94} & \textbf{86.17} & \textbf{55.83} & \textbf{76.12} \\
\hline

\hline
\end{tabular}
\caption{\label{Performance-2}
Compared to Prototypical-ADV which uses additional unlabeled target domain instances, our MBATF model which does not use any extra target-domain instances performs similarly on FewRel 2.0 and Wiki$\to$Semeval tasks, and even better on FewRel 1.0 and Wiki$\to$NYT tasks.
}
\end{table*}

\section{Conclusion}
In this paper, we propose novel MBATF composed of attention-based Prototypical Networks and Meta-ADV for few-shot cross-domain RC tasks. Our contributions are mainly in three aspects: realizing ADV without extra target-domain data; proposing a new idea of ADV for few-shot and cross-domain tasks; adopting attention mechanism to find common features between different fields. In the experiments, we test our model on various cross-domain issues, including FewRel 2.0, FewRel 1.0, and two new set-ups constructed by ourselves, which demonstrates that our model can still achieve desirable results without target-domain data. In the future, we will further explore the application of our Meta-ADV module to other metric-based meta-learning to make it more comprehensive.

\bibliography{anthology, anthology_extra, acl2020}
\bibliographystyle{acl_natbib}

\end{document}


\maketitle
  \appendix
  \renewcommand{\appendixname}{Appendix~\Alph{section}}

  \section{Computing Infrastructure}
  All our experiments runs on Tesla V100-PCIE-32GB, Quadro RTX 6000, and TITAN Xp. 
  
  \section{Performance on Validation set}
  The Performance on Validation set is shown in Table \ref{tb:fewrel}.
  \begin{table*}
  \centering
  \begin{tabular}{|c|c|c|c|c|c|}
  \hline
  \multirow{2}{*}{Datasets} &
  \multirow{2}{*}{Model} &
  \multirow{2}{*}{5 way 1 shot} &
  \multirow{2}{*}{5 way 5 shot} &
  \multirow{2}{*}{10 way 1 shot} &
  \multirow{2}{*}{10 way 5 shot} 
  \\
  & & & & & \\
  
  \hline
  \multirow{2}{*}{\shortstack{FewRel 2.0\\ (Wiki$\to$Pubmed)}}
  & {ProtoNets} & {36.85} & {46.17} & {20.48} & {33.85}  \\
  & {MBATF} & \textbf{39.50} & \textbf{56.69} & \textbf{28.35} & \textbf{42.45}  \\
  \hline
  \multirow{2}{*}{\shortstack{FewRel 1.0\\ (Wiki$\to$Wiki)}}
  & {ProtoNets} & {70.69} & {84.20} & {58.53} & {74.71} \\
  & {MBATF} & \textbf{71.42} & \textbf{86.44} & \textbf{59.79} & \textbf{76.15} \\
  \hline
  \end{tabular}
  \caption{\label{tb:fewrel} Performance of our MBATF (ProtoNets+MetaADV+RelationScore) on Validation sets}
  \end{table*}
  
  \section{Relevant statistics}
  As for FewRel 2.0, it contains 64-class Wikidata in meta-training, each class has 700 instances and 25-class Pubmed data in meta-testing, each class has 100 instances.
  
  As for FewRel 1.0, it contains 64-class Wikidata in meta-training, and 20-class Wikidata in meta-testing. Each class also involves 700 instances.
  
  As for SemEval-2010 Task 8, it contains 9 classes and totally 6674 instances.
  
  As for NYT-10, it contains 57 classes and totally 143391 instances.